\documentclass{article} 
\usepackage{nips15submit_e,times}
\usepackage{hyperref}
\usepackage{url}
\usepackage[numbers,sort]{natbib}
\usepackage{graphicx}
\usepackage{amsmath,amssymb}
\usepackage{array}
\usepackage{multirow}
\usepackage{color,colortbl}
\usepackage[compact]{titlesec}
\usepackage{array}
\usepackage{enumitem}
\usepackage{appendix}

\nipsfinalcopy 

\setenumerate{noitemsep,topsep=0pt,parsep=0pt,partopsep=0pt,labelindent=0pt,
leftmargin=1.2em}
\titlespacing{\section}{0pt}{0pt}{0pt}
\titlespacing{\subsection}{0pt}{0pt}{0pt}
\newcommand{\figref}[1]{Fig.~\ref{#1}}
\newcommand{\tblref}[1]{Table~\ref{#1}}
\newcommand{\sref}[1]{Sect.~\ref{#1}}

\def\eg{\emph{e.g.}}

\def\etal{\emph{et al.}}

\definecolor{Gray}{gray}{0.95}

\def\mat#1{\mathchoice{\mbox{\boldmath$\displaystyle\tt#1$}}
{\mbox{\boldmath$\textstyle\tt#1$}}
{\mbox{\boldmath$\scriptstyle\tt#1$}}
{\mbox{\boldmath$\scriptscriptstyle\tt#1$}}}
\def\m#1{\protect\mat #1}

\renewcommand{\paragraph}[1]{\par\noindent{\bf #1}}

\begin{document}
\title{Spatial Transformer Networks}
\author{
Max Jaderberg \\
\And
Karen Simonyan \\
\And
Andrew Zisserman \\
\And
Koray Kavukcuoglu
}
\maketitle
 \vspace{-3.0em}
 \begin{center}
 Google DeepMind, London, UK\\
 \texttt{\{jaderberg,simonyan,zisserman,korayk\}@google.com}
 \end{center}
 \vspace{1em}
\begin{abstract}
Convolutional Neural Networks define an exceptionally
powerful class of models, but are still limited by the lack of ability
to be spatially invariant to the input data in a computationally and parameter
efficient manner. In this work we introduce a new learnable module, the
\emph{Spatial Transformer}, which explicitly allows the spatial manipulation of
data within the network. This differentiable module can be inserted
into existing convolutional architectures, giving neural networks the ability to
actively spatially transform feature maps, conditional on the feature map itself,
without any extra training supervision or modification to the optimisation process. We show that the use
of spatial transformers results in models which learn invariance to translation,
scale, rotation and more generic warping, resulting in state-of-the-art
performance on several benchmarks, and for a number
of classes of transformations.
\end{abstract}

\section{Introduction}

Over recent years, the landscape of computer vision has been
drastically altered and pushed forward through the adoption of a fast,
scalable, end-to-end learning framework, the Convolutional Neural
Network (CNN)~\cite{Lecun98}. Though not a recent invention, we now
see a cornucopia of CNN-based models achieving state-of-the-art results
in classification~\cite{Szegedy14,Jaderberg14c,Schroff15},
localisation~\cite{Simonyan14c,Tompson14}, semantic
segmentation~\cite{Long15}, and action
recognition~\cite{Simonyan14video,Gkioxari15} tasks, amongst others.


A desirable property of a system which is able to reason about images is to disentangle
object pose and part deformation from texture and shape. The introduction of
local max-pooling layers in CNNs has helped to satisfy this property by
allowing a network to be somewhat spatially invariant to the
position of features. However, due to the typically small spatial
support for max-pooling (\eg~$2 \times 2$ pixels) this spatial
invariance is only realised over a deep hierarchy of max-pooling and
convolutions, and the intermediate feature maps (convolutional layer activations)
in a CNN are not actually invariant to large transformations of the input
data~\cite{Cohen15,Lenc14}. This limitation of CNNs is due to having
only a limited, pre-defined pooling mechanism for dealing with variations in the spatial
arrangement of data.

In this work we introduce a {\em Spatial Transformer} module, that can
be included into a standard neural network architecture to provide
spatial transformation capabilities. The action of the spatial transformer is
conditioned on individual data samples, with the appropriate behaviour learnt
during training for the task in question (without extra supervision).
Unlike pooling layers, where the
receptive fields are fixed and local, the spatial transformer module is a dynamic
mechanism that can actively spatially transform an image
(or a feature map) by producing an appropriate transformation for each input sample.
The transformation is then performed on the entire feature map (non-locally) and
can include scaling, cropping, rotations, as well as non-rigid deformations.
This allows networks which include spatial
transformers to not only select regions of an image
that are most relevant (attention), but also to transform those regions
to a canonical, expected pose to simplify recognition in the following layers.
Notably, spatial transformers can be trained with standard back-propagation,
allowing for end-to-end training of the models they are injected in.

\begin{figure}[t]
\raisebox{-\height}{\vspace{0pt}\includegraphics[width=0.42\textwidth]{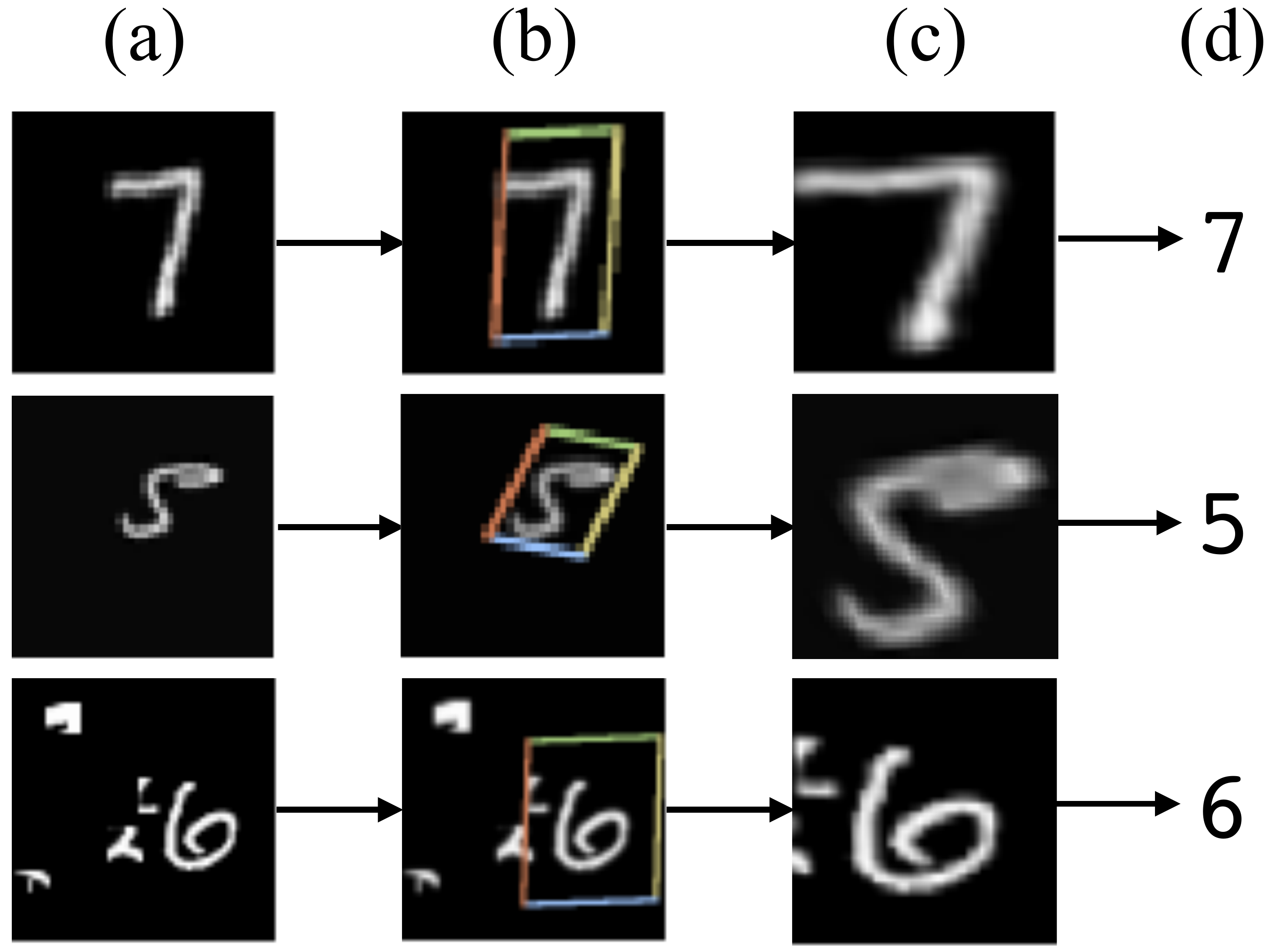}}\hfill%
\begin{minipage}[t]{0.55\textwidth}\vspace{-1em}%
\caption{\small The result of using a spatial transformer as the first layer of
a fully-connected network trained for distorted MNIST digit classification. (a)
The input to the spatial transformer network is an image of
an MNIST digit that is distorted with random translation, scale, rotation, and clutter. (b) The
localisation network of the spatial transformer predicts a transformation to
apply to the input image. (c) The output of the spatial transformer, after
applying the transformation. (d) The classification prediction
produced by the subsequent fully-connected network on the output of the spatial
transformer. The spatial transformer network (a CNN including a spatial transformer
module) is trained end-to-end with only class labels -- no knowledge of the
groundtruth transformations is given to the system.}
\label{fig:teaser}
\end{minipage}
\end{figure}

Spatial transformers can be incorporated into CNNs to benefit
multifarious tasks, for example:
(i)~{\em image classification:}
suppose a CNN is trained to perform multi-way classification of images
according to whether they contain a particular digit -- where the
position and size of the digit may vary significantly with each sample
(and are uncorrelated with the class); a spatial transformer that
crops out and scale-normalizes the appropriate region can simplify the
subsequent classification task, and lead to superior
classification performance, see \figref{fig:teaser};
(ii)~{\em co-localisation:}
given a set of images containing different instances of the same (but unknown)
class, a spatial transformer can be used to localise them in each image;
(iii)~{\em spatial attention:} a spatial transformer can be used for tasks requiring an attention mechanism, such as in \cite{Gregor15,Xu15}, but is more flexible and can be trained purely with backpropagation without reinforcement learning. A key benefit of using attention is that transformed (and so attended), lower resolution inputs can be used in favour of higher resolution raw inputs, resulting in increased computational efficiency.

The rest of the paper is organised as follows: \sref{sec:related} discusses some work related to our own, we introduce the formulation and implementation of the spatial transformer in \sref{sec:transformers}, and finally give the results of experiments in \sref{sec:experiments}. Additional experiments and implementation details are given in Appendix~\ref{sec:morexp}.

\section{Related Work}\label{sec:related}
In this section we discuss the prior work related to the paper, covering the central
ideas of modelling transformations with neural networks~\cite{Hinton81,Hinton11,Tieleman14},
learning and analysing transformation-invariant
representations~\cite{Gens14,Cohen15,Sohn12,Kanazawa14,Bruna13,Lenc14}, as well
as attention and detection mechanisms for feature
selection~\cite{Schmidhuber91,Ba14,Sermanet14b,Gregor15,Girshick14,Erhan14}.

Early work by Hinton~\cite{Hinton81} looked at assigning canonical frames of reference to object parts, a theme which recurred in~\cite{Hinton11} where 2D
affine transformations were modeled to create a generative model composed of transformed
parts. The targets of the generative training scheme are the transformed input
images, with the transformations between input images and targets given as an
additional input to the network. The result is a generative model which can
learn to generate transformed images of objects by composing parts. The notion
of a composition of transformed parts is taken further by Tieleman~\cite{Tieleman14},
where learnt parts are explicitly affine-transformed, with the transform predicted
by the network. Such generative capsule models are able to learn discriminative
features for classification from transformation supervision.

The invariance and equivariance of CNN representations to input image
transformations are studied in \cite{Lenc14} by estimating the linear relationships
between representations of the original and transformed images.
Cohen \& Welling~\cite{Cohen15} analyse this behaviour in relation to symmetry groups,
which is also exploited in the architecture proposed by Gens \& Domingos~\cite{Gens14},
resulting in feature maps that are more invariant to symmetry groups. Other attempts
to design transformation invariant representations are scattering networks~\cite{Bruna13},
and CNNs that construct filter banks of transformed filters~\cite{Sohn12,Kanazawa14}.
Stollenga~\etal~\cite{Stollenga14} use a policy based on a network's
activations to gate the responses of the network's filters for a subsequent
forward pass of the same image and so can allow attention to specific features.
In this work, we aim to achieve invariant representations by manipulating the data
rather than the feature extractors, something that was done for clustering in~\cite{Frey01b}.

Neural networks with selective attention manipulate the data by taking crops, and so
are able to learn translation invariance. Work such as~\cite{Ba14,Sermanet14b}
are trained with reinforcement learning to avoid the need for a differentiable attention
mechanism, while~\cite{Gregor15} use a differentiable attention mechansim by utilising
Gaussian kernels in a generative model. The work by Girshick~\etal~\cite{Girshick14}
uses a region proposal algorithm as a form of attention, and~\cite{Erhan14} show that
it is possible to regress salient regions with a CNN. The framework we present in this
paper can be seen as a generalisation of differentiable attention to any spatial transformation.

\section{Spatial Transformers}\label{sec:transformers}
\begin{figure}[t]
\centering
\begin{tabular}{cccc}
\includegraphics[width=0.75\textwidth]{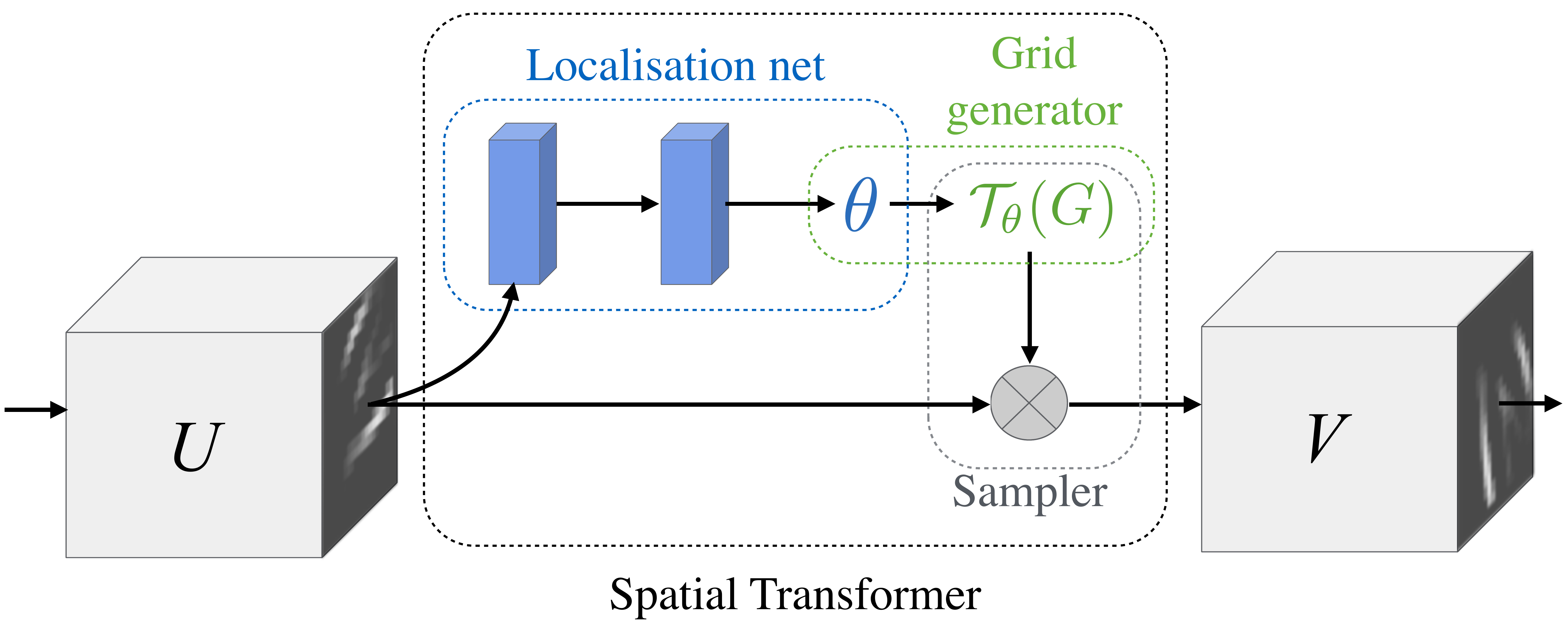}
\end{tabular}
\caption{\small The architecture of a spatial transformer module. The input
feature map $U$ is passed to a localisation network which regresses the
transformation parameters $\theta$. The regular spatial grid $G$ over $V$ is transformed to the
sampling grid ${\cal T}_\theta (G)$, which is applied to $U$ 
as described in \sref{sec:gridsampler}, producing the warped
output feature map $V$. The combination of the localisation network and sampling
mechanism defines a spatial transformer.}
\label{fig:stnet}
\end{figure}

In this section we describe the formulation of a \emph{spatial transformer}.
This is a differentiable module which applies a spatial transformation to a feature map during a single forward pass, where the transformation is
conditioned on the particular input, producing a single output feature map. For multi-channel inputs, the same warping is applied to each channel. For simplicity, in this section we consider single transforms and single outputs per transformer, however we can generalise to multiple transformations, as shown in experiments.

The spatial transformer mechanism is split into three parts, shown in
\figref{fig:stnet}. In order of computation, first a \emph{localisation network}
(\sref{sec:locnet}) takes the input feature map, and through a number of hidden
layers outputs the parameters of the spatial transformation that should be applied
to the feature map -- this gives a transformation conditional on the
input. Then, the predicted transformation parameters are used
to create a sampling grid, which is a set of points where the input map should be
sampled to produce the transformed output.
This is done by the \emph{grid generator}, described in \sref{sec:affinegrid}.
Finally, the feature map and the sampling grid
are taken as inputs to the \emph{sampler}, producing the output map
sampled from the input at the grid points (\sref{sec:gridsampler}).

The combination of these three components forms a spatial transformer and
will now be described in more detail in the following sections.

\subsection{Localisation Network}\label{sec:locnet}

The localisation network takes the input feature map $U \in \mathbb{R}^{H \times
W \times C}$ with width $W$, height $H$ and $C$ channels and outputs $\theta$,
the parameters of the transformation ${\cal T}_\theta$ to be applied to the
feature map: $\theta = f_\text{loc}(U)$. The size of $\theta$ can vary depending
on the transformation type that is parameterised, \eg~for an affine
transformation $\theta$ is 6-dimensional as in \eqref{eqn:affine}.

The localisation network function $f_\text{loc}()$ can take any form, such as a
fully-connected network or a convolutional network, but should include a final
regression layer to produce the transformation parameters $\theta$.

\subsection{Parameterised Sampling Grid}\label{sec:affinegrid}

To perform a warping of the input feature map, each output pixel is computed by applying a sampling kernel centered at a particular
location in the input feature map (this is described fully in the next
section). By \emph{pixel} we refer to an element of a generic feature map, not necessarily an image. In general, the output pixels are defined to lie on a regular
grid $G = \{G_i\}$ of pixels $G_{i} =
(x_i^t, y_i^t)$, forming an output feature map
$V \in \mathbb{R}^{H' \times W' \times C}$, where
$H'$ and $W'$ are the height and width of the grid, and $C$ is the number of
channels, which is the same in the input and output.

For clarity of exposition, assume for the moment that ${\cal T}_\theta$ is a
2D affine transformation $\m A_\theta$. We will discuss other transformations below.
In this affine case, the pointwise transformation is
\begin{equation}
\left( {
\renewcommand{\arraystretch}{1.2}
\begin{array}{c}
x_i^s\\
y_i^s
\end{array}
}\right)
= {\cal T}_\theta (G_i)
= \m A_\theta
\left( {
\renewcommand{\arraystretch}{1.2}
\begin{array}{c}
x^t_i\\
y^t_i\\
1
\end{array}
}\right)
=
\left[ {
\renewcommand{\arraystretch}{1.2}
\begin{array}{ccc}
\theta_{11} & \theta_{12} & \theta_{13}\\
\theta_{21} & \theta_{22} & \theta_{23}
\end{array}
}\right]
\left( {
\renewcommand{\arraystretch}{1.2}
\begin{array}{c}
x^t_i\\
y^t_i\\
1
\end{array}
}\right)
\label{eqn:affine}
\end{equation}
where $(x_i^t, y_i^t)$ are the target coordinates of the regular grid
in the output feature map, $(x_i^s,y_i^s)$ are the source
coordinates in the input feature map that define the sample points, and $\m A_\theta$
is the affine transformation matrix. We use height and width normalised
coordinates, such that $-1 \le x_i^t, y_i^t \le 1$ when within the spatial bounds of
the output, and $-1 \le x_i^s, y_i^s \le 1$ when within the spatial bounds of the
input (and similarly for the $y$ coordinates).
The source/target transformation and sampling is equivalent to the
standard texture mapping and coordinates used in graphics~\cite{Foley94}.

\begin{figure}[t]
\centering
\begin{tabular}{cccc}
\includegraphics[width=0.3\textwidth]{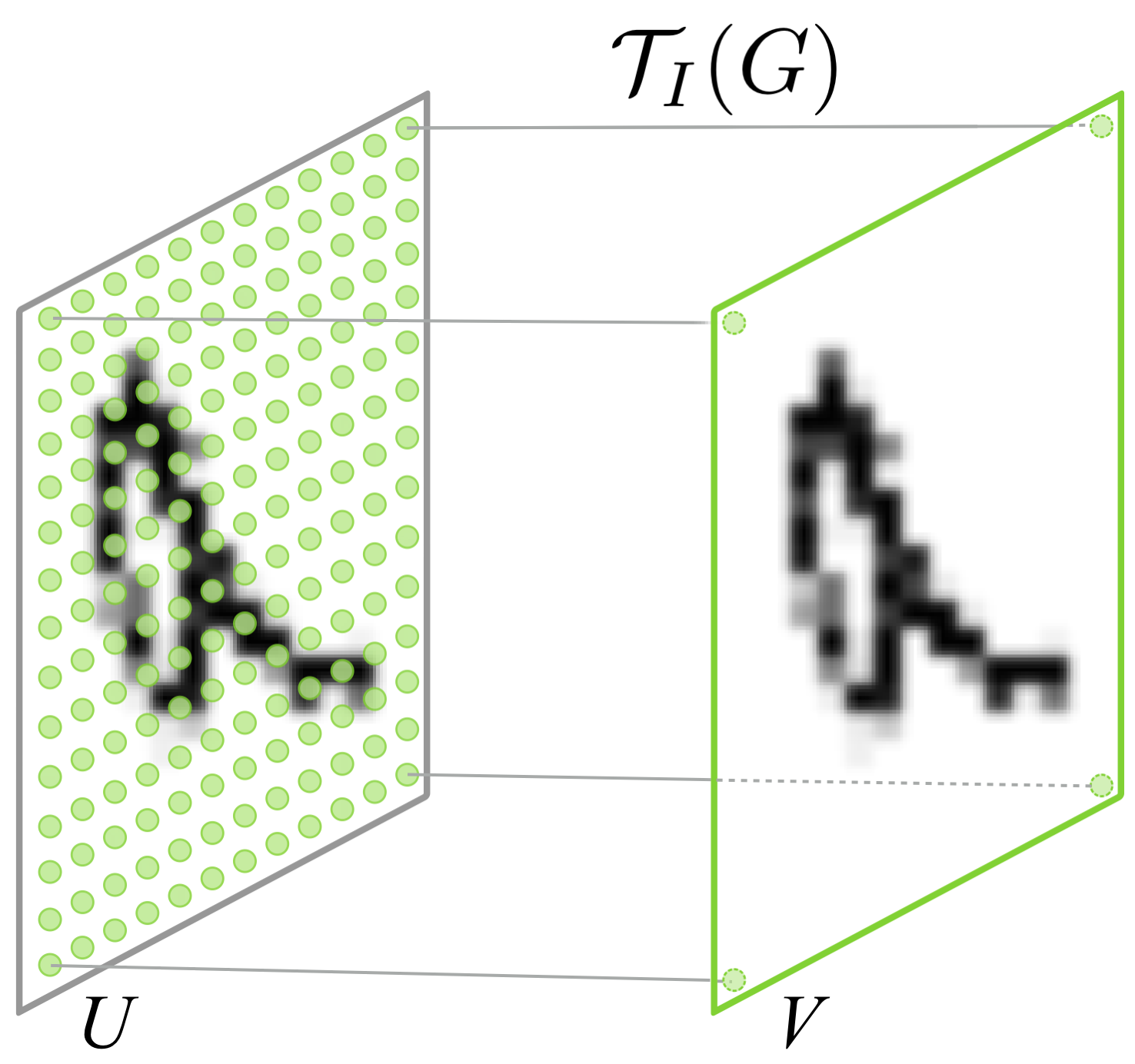}~~~~~~~~~~~~~~~~~~&
\includegraphics[width=0.3\textwidth]{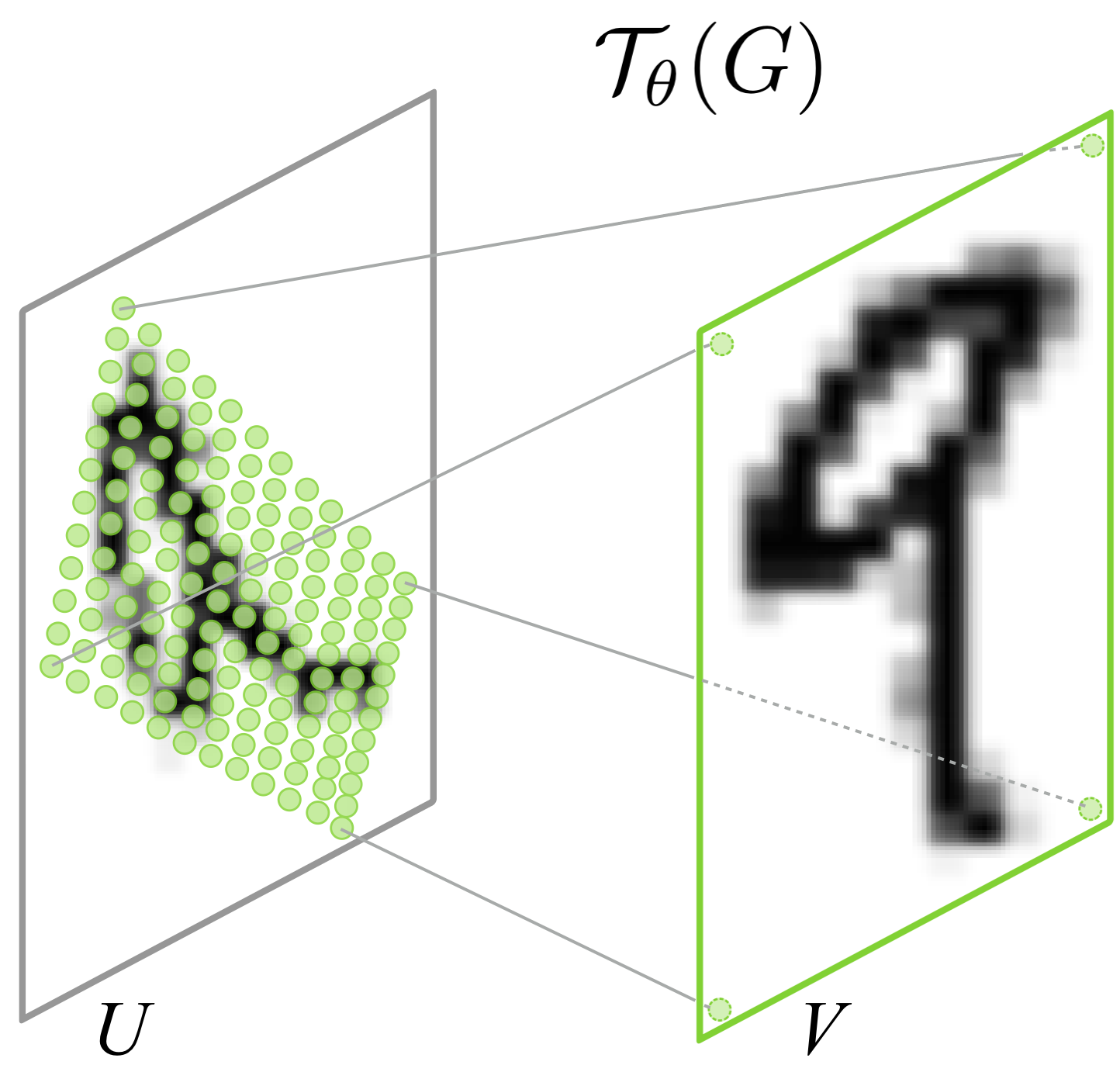}
\\
(a)&(b)
\end{tabular}
\caption{\small Two examples of applying the parameterised sampling grid to an
image $U$ producing the output $V$. (a) The sampling grid is the regular grid $G
= {\cal T}_I (G)$, where $I$ is the identity transformation parameters. (b) The
sampling grid is the result of warping the regular grid with an affine
transformation ${\cal T}_\theta (G)$.}
\label{fig:grid}
\end{figure}

The transform defined in \eqref{eqn:affine} allows cropping,
translation, rotation, scale, and skew to be applied to the input
feature map, and requires only 6 parameters (the 6 elements of $\m A_\theta$) to
be produced by the localisation network. It allows cropping because
if the transformation is a contraction (i.e.\ the determinant of
the left $2 \times 2$ sub-matrix has magnitude less than unity)
then the mapped regular grid will lie in a parallelogram of area
less than the range of $x_i^s,y_i^s$. The effect of this
transformation on the grid compared to the identity transform is shown
in \figref{fig:grid}.

The class of transformations ${\cal T}_\theta$ may be
more
constrained, such as that used for attention
\begin{equation}
\m A_\theta=
\left[ {
\begin{array}{ccc}
s & 0 & t_x\\
0 & s & t_y
\end{array}
}\right]
\label{eqn:attention}
\end{equation}
allowing cropping, translation, and isotropic scaling by varying $s$, $t_x$, and
$t_y$. The transformation  ${\cal T}_\theta$ can also be more general, such
as a plane projective transformation with 8 parameters, piecewise affine,
or a thin plate spline. Indeed, the transformation can have any parameterised
form,
provided that it is
differentiable with respect to the parameters -- this crucially allows gradients
to be backpropagated through from the sample points ${\cal T}_\theta (G_i)$
to the localisation
network output $\theta$.
If the transformation is parameterised
in a structured, low-dimensional way, this
reduces the complexity of the task assigned to the localisation
network. For instance, a generic class of structured and differentiable
transformations, which is a superset of attention, affine, 
projective, and thin plate spline transformations, is ${\cal T}_\theta = M_\theta
B$, where $B$ is a target grid representation (\eg~in \eqref{eqn:affine},
$B$ is the regular grid $G$ in homogeneous coordinates), and $M_\theta$ is a matrix parameterised by $\theta$.
In this case it is possible to not only
learn how to predict $\theta$ for a sample, but also to learn $B$ for the task
at hand.

\subsection{Differentiable Image Sampling}\label{sec:gridsampler}
To perform a spatial transformation of the input feature map, a sampler must take
the set of sampling points ${\cal T}_\theta (G)$, along with the input
feature map $U$ and produce the sampled output feature map $V$.

Each $(x^s_i,y^s_i)$ coordinate in ${\cal T}_\theta (G)$ defines the spatial
location in the input where a sampling kernel is applied to get the
value at a particular pixel in the output $V$. This can be written as
\begin{equation}
V_i^c = \sum_{n}^{H} \sum_{m}^W U^c_{nm}k(x_i^s - m;
\Phi_x)k(y_i^s-n;\Phi_y)~~\forall i\in[1\ldots H'W']~ ~\forall c\in[1\ldots C]
\label{eqn:sampling}
\end{equation}
where $\Phi_x$ and $\Phi_y$ are the parameters of a generic sampling kernel
$k()$ which defines the image interpolation (\eg~bilinear), $U^c_{nm}$ is the
value at location $(n,m)$ in channel $c$ of the input, and $V_i^c$ is
the output value for pixel $i$ at location $(x_i^t, y_i^t)$ in channel $c$. Note
that the sampling is done identically for each channel of the input, so every
channel is transformed in an identical way (this preserves spatial consistency
between channels).

In theory, any sampling kernel can be used, as long as (sub-)gradients can be defined
with respect to $x_i^s$ and $y_i^s$. For example, using the integer sampling
kernel reduces \eqref{eqn:sampling} to
\begin{equation}
V_i^c = \sum_{n}^{H} \sum_{m}^W U^c_{nm}\delta(\lfloor x_i^s + 0.5 \rfloor -
m)\delta(\lfloor y_i^s + 0.5 \rfloor - n)
\label{eqn:integer}
\end{equation}
where $\lfloor  x + 0.5 \rfloor$ rounds $x$ to the nearest integer and
$\delta()$ is the Kronecker delta function. This sampling kernel equates to just
copying the value at the nearest pixel to $(x_i^s,y_i^s)$ to the output location
$(x_i^t,y_i^t)$. Alternatively, a bilinear sampling kernel can be used, giving
\begin{equation}
V_i^c = \sum_{n}^{H} \sum_{m}^W U^c_{nm}\max(0, 1-|x_i^s-m|)\max(0,1-|y_i^s-n|)
\label{eqn:bilinear}
\end{equation}

To allow backpropagation of the loss through this sampling mechanism we can
define the gradients with respect to $U$ and $G$. For bilinear sampling
\eqref{eqn:bilinear} the partial derivatives are
\begin{equation}
\frac{\partial V^c_i}{\partial U^c_{nm}} =
\sum_{n}^{H} \sum_{m}^W \max(0, 1-|x_i^s-m|)\max(0,1-|y_i^s-n|)
\label{eqn:dvbydu}
\end{equation}
\begin{equation}
\frac{\partial V^c_i}{\partial x_i^s} =
\sum_{n}^{H} \sum_{m}^W U^c_{nm}\max(0,1-|y_i^s-n|)
\begin{cases}
		0  & \mbox{if}~ |m - x_i^s| \ge 1 \\
		1  & \mbox{if}~ m \ge x_i^s\\
		-1  & \mbox{if}~ m < x_i^s\\
\end{cases}
\label{eqn:dvbydx}
\end{equation}
and similarly to \eqref{eqn:dvbydx} for $\frac{\partial V^c_i}{\partial
y_i^s}$.

This gives us a (sub-)differentiable sampling mechanism, allowing loss gradients to
flow back not only to the input feature map~\eqref{eqn:dvbydu}, but also to the sampling grid
coordinates~\eqref{eqn:dvbydx}, and therefore back to the transformation parameters $\theta$ and
localisation network since $\frac{\partial x_i^s}{\partial \theta}$ and
$\frac{\partial x_i^s}{\partial \theta}$ can be easily derived from
\eqref{eqn:affine} for example. Due to discontinuities in the sampling fuctions,
sub-gradients must be used. This sampling mechanism can be implemented very
efficiently on GPU, by ignoring the sum over all input locations and instead just
looking at the kernel support region for each output pixel.

\subsection{Spatial Transformer Networks}
The combination of the localisation network, grid generator, and sampler form a
spatial transformer (\figref{fig:stnet}). This is a self-contained module which can
be dropped into a CNN architecture at any point, and in any number, giving rise
to \emph{spatial transformer networks}. This module is computationally very fast
and does not impair the training speed, causing very
little time overhead when used naively, and even speedups in attentive models due to subsequent downsampling that can be applied to the output of the transformer.

Placing spatial transformers within a CNN allows the network to learn how to
actively transform the feature maps to help minimise the overall cost function of
the network during training. The knowledge of how to transform each training
sample is compressed and cached in the weights of the localisation network (and
also the weights of the layers previous to a spatial transformer) during
training. For some
tasks, it may also be useful to feed the output of the localisation network,
$\theta$, forward to the rest of the network, as it explicitly encodes the
transformation, and hence the pose, of a region or object.

It is also possible to use spatial transformers to downsample or oversample a
feature map, as one can define the output dimensions $H'$ and $W'$ to be
different to the input dimensions $H$ and $W$. However, with sampling kernels
with a fixed, small spatial support (such as the bilinear kernel), downsampling with a
spatial transformer can cause aliasing effects.


Finally, it is possible to have multiple spatial transformers in a CNN. Placing multiple spatial transformers at increasing depths of a network allow transformations of increasingly abstract representations, and also gives the localisation networks potentially more informative representations to base the predicted transformation parameters on. One can also use multiple spatial transformers
in parallel -- this can be useful if there are multiple objects or parts of
interest in a feature map that should be focussed on individually.
A limitation of this architecture in a purely feed-forward network is that the
number of parallel spatial transformers limits the number of objects that the
network can model.

\section{Experiments}\label{sec:experiments}
In this section we explore the use of spatial transformer networks on a number
of supervised learning tasks. In \sref{sec:mnist} we begin with experiments on
distorted versions of the MNIST handwriting dataset, showing the
ability of spatial transformers to improve classification performance through
actively transforming the input images. In \sref{sec:svhn} we test spatial
transformer networks on a challenging real-world dataset, Street View House
Numbers~\cite{Netzer11}, for number recognition, showing state-of-the-art
results using multiple spatial transformers embedded in the convolutional stack
of a CNN. Finally, in \sref{sec:fgc}, we investigate the use of multiple
parallel spatial transformers for fine-grained classification, showing
state-of-the-art performance on CUB-200-2011 birds dataset~\cite{Wah11} by discovering object parts and learning to attend to them. Further experiments of MNIST addition and co-localisation can be found in Appendix~\ref{sec:morexp}.

\subsection{Distorted MNIST}\label{sec:mnist}
\begin{table}[t]
\begin{center}\small
\setlength{\tabcolsep}{2pt}
\begin{tabular}[t]{ll||c|c|c|c}
\multicolumn{2}{p{1.5cm}||}{~}&
\multicolumn{4}{c}{\centering MNIST Distortion}\\
\multicolumn{2}{p{1.5cm}||}{\centering ~~~~Model} &
\multicolumn{1}{c|}{\centering R} &
\multicolumn{1}{c|}{\centering RTS} &
\multicolumn{1}{c|}{\centering P} &
\multicolumn{1}{c}{\centering E}\\
\hline
\multicolumn{2}{l||}{FCN} & 2.1 & 5.2 & 3.1 & 3.2\\
\multicolumn{2}{l||}{CNN} & 1.2 & 0.8 & 1.5 & 1.4\\
\hline
\multirow{3}{*}{ST-FCN}
    & Aff & 1.2 & 0.8 & 1.5 & 2.7\\
    & Proj & 1.3 & 0.9 & 1.4 & 2.6\\
    & TPS & 1.1 & 0.8 & 1.4 & 2.4\\
\hline
\multirow{3}{*}{ST-CNN}
    & Aff & 0.7 & 0.5 & 0.8 & 1.2\\
    & Proj & 0.8 & 0.6 & 0.8 & 1.3\\
    & TPS & 0.7 & 0.5 & 0.8 & 1.1\\
\end{tabular}
\qquad\quad
\hspace{-3em}
\vtop{\vspace{-1em}\hbox{\includegraphics[width=0.64\textwidth]{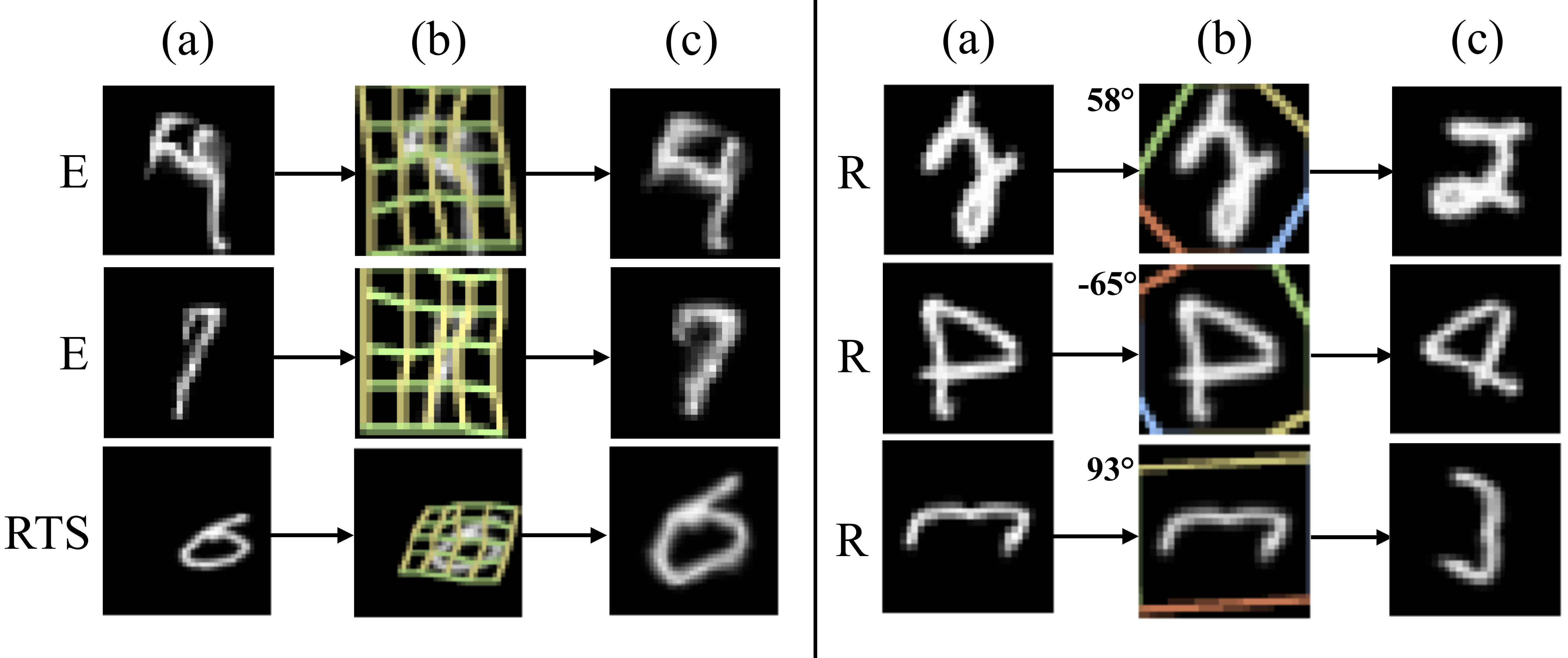}}}
\end{center}
\caption{\small \emph{Left:} The percentage errors for different models on different
distorted MNIST datasets. The different distorted MNIST datasets we test are TC: translated and cluttered, R: rotated, RTS:
rotated, translated, and scaled, P: projective distortion, E: elastic
distortion. All the models used for each experiment have the same number of
parameters, and same base structure for all experiments. \emph{Right:} Some
example test images where a spatial transformer network correctly classifies the
digit but a CNN fails. 
(a) The inputs to the networks. 
(b) The transformations predicted by the spatial transformers, visualised by the
grid $T_{\theta}(G)$.
(c) The outputs of the spatial transformers.
E and RTS examples use thin plate spline spatial transformers (ST-CNN TPS),
while R examples use affine spatial transformers (ST-CNN Aff) with the angles of the affine transformations given. For videos showing animations of these experiments and more see \url{https://goo.gl/qdEhUu}.}
\label{table:mnist}
\end{table}

In this section we use the MNIST handwriting dataset as a
testbed for exploring the range of transformations to which a network can learn invariance to by using a spatial transformer.

We begin with experiments where we train different neural network models to classify
MNIST data that has been distorted in various ways: rotation (R),
rotation, scale and translation (RTS),
projective transformation (P),
and elastic warping (E) -- note that elastic warping is destructive and can not be inverted in some cases. 
The full details of the distortions used to generate this data are given in Appendix~\ref{sec:morexp}.
We train baseline fully-connected (FCN) and convolutional (CNN) neural networks,
as well as networks with spatial transformers acting on the input before the
classification network (ST-FCN and ST-CNN). The spatial transformer networks all
use bilinear sampling, but variants use different transformation functions: an
affine transformation (Aff), projective transformation (Proj), and a 16-point
thin plate spline transformation (TPS)~\cite{Bookstein89}. The CNN models include two max-pooling layers. All networks have approximately the same
number of parameters, are trained with identical optimisation schemes
(backpropagation, SGD, scheduled learning rate decrease, with a multinomial
cross entropy loss), and all with three weight layers in the
classification network.

The results of these experiments are shown in \tblref{table:mnist} (left).
Looking at any particular type of distortion of the data, it is clear that a
spatial transformer enabled network outperforms its counterpart base network.
For the case of rotation, translation, and scale distortion (RTS), the ST-CNN
achieves 0.5\% and 0.6\% depending on the class of transform used for ${\cal
T}_\theta$, whereas a CNN, with two max-pooling layers to provide spatial
invariance, achieves 0.8\% error. This is in fact the same error that the ST-FCN
achieves, which is without a single convolution or max-pooling layer in its
network, showing that using a spatial transformer is an alternative way to achieve spatial invariance. ST-CNN models consistently perform better than ST-FCN models due to max-pooling 
layers in ST-CNN providing even more spatial invariance, and convolutional layers better modelling local structure. We also test our models in a noisy environment, on $60 \times 60$ images with translated MNIST digits and background clutter (see \figref{fig:teaser} third row for an example): an FCN gets 13.2\% error, a CNN gets 3.5\% error, while an ST-FCN gets 2.0\% error and an ST-CNN gets 1.7\% error.

Looking at the results between different classes of transformation, the
thin plate spline transformation (TPS) is the most powerful, being able to
reduce error on elastically deformed digits by reshaping the input into a
prototype instance of the digit, reducing the complexity of the task for the
classification network, and does not over fit on simpler data \eg~R. Interestingly, the transformation of inputs for all ST
models leads to a ``standard'' upright posed digit -- this is the mean pose found in the training data. In \tblref{table:mnist} (right), we show the transformations performed for some test cases where a CNN is unable to correctly classify the digit, but a
spatial transformer network can. Further test examples are visualised in an animation here \url{https://goo.gl/qdEhUu}.

\subsection{Street View House Numbers}\label{sec:svhn}

\begin{table}[t]
\begin{center}\small
\setlength{\tabcolsep}{3pt}
\begin{tabular}[t]{ll||c|c}
\\
\multicolumn{2}{p{2.0cm}||}{~}&
\multicolumn{2}{c}{\centering Size}\\
\multicolumn{2}{p{2.0cm}||}{\centering ~~~~~~Model} &
\multicolumn{1}{c|}{\centering 64px} &
\multicolumn{1}{c}{\centering 128px}\\
\hline
\multicolumn{2}{l||}{Maxout CNN~\cite{Goodfellow13}} & 4.0 & -\\
\multicolumn{2}{l||}{CNN (ours)} & 4.0 & 5.6 \\
\multicolumn{2}{l||}{DRAM\textsuperscript{*}~\cite{Ba14}} & 3.9 & 4.5 \\
\hline
\multirow{2}{*}{ST-CNN}
    & Single & 3.7 & \bf{3.9} \\
    & Multi & \bf{3.6} & \bf{3.9} \\
\end{tabular}
\qquad\quad
\hspace{-2em}
\vtop{\vspace{-1em}\hbox{\includegraphics[width=0.6\textwidth]{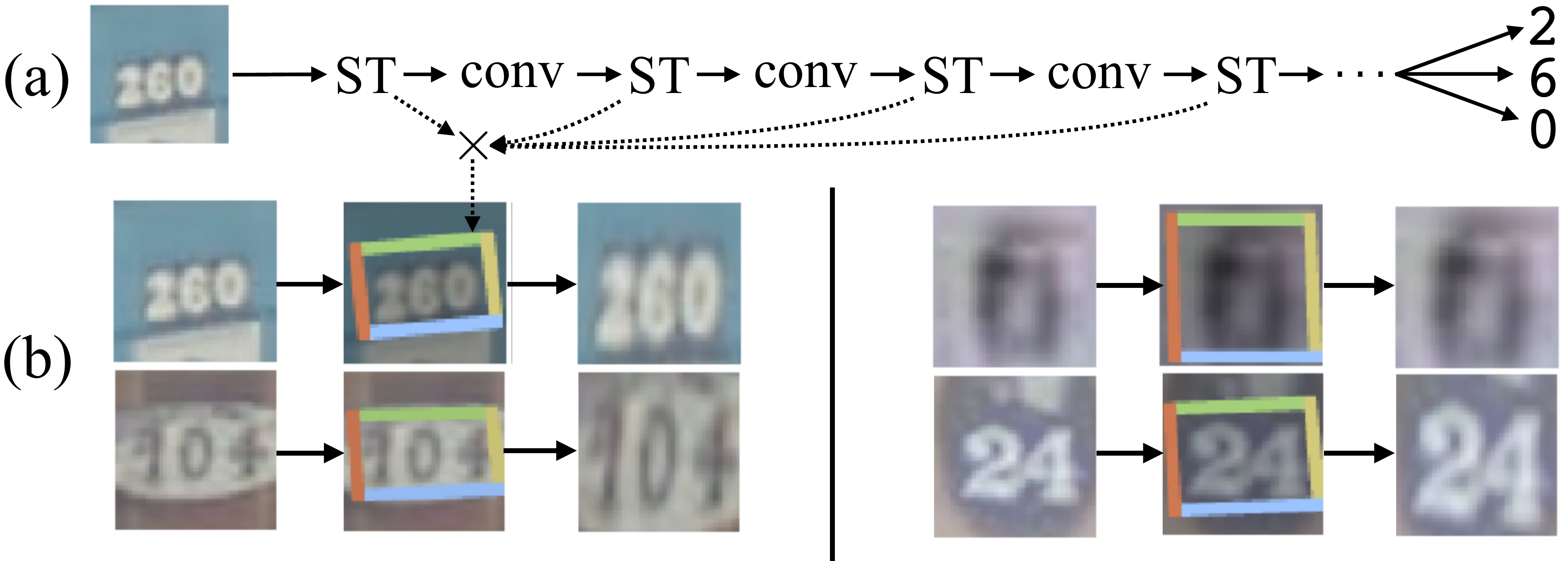}}}%
\end{center}
\vspace{-1em}
\caption{\small \emph{Left:} The sequence error for SVHN multi-digit recognition
on crops of $64\times 64$ pixels (64px), and inflated crops of $128 \times 128$
(128px) which include more background. \textsuperscript{*}The best reported
result from \cite{Ba14} uses model averaging and Monte Carlo averaging, whereas
the results from other models are from a single forward pass of a single model.
\emph{Right:} (a) The schematic of the ST-CNN Multi model. The transformations
applied by each spatial transformer (ST) is applied to the convolutional feature map produced by the previous layer. (b)
The result of multiplying out the affine transformations predicted by the four
spatial transformers in ST-CNN Multi, visualised on the input image.}
\label{table:svhn}
\end{table}

We now test our spatial transformer networks on a challenging real-world
dataset, Street View House Numbers (SVHN)~\cite{Netzer11}. This dataset contains
around 200k real world images of house numbers, with the task to recognise the
sequence of numbers in each image. There are between 1 and 5 digits in each
image, with a large variability in scale and spatial arrangement.

We follow the experimental setup as in \cite{Goodfellow13,Ba14}, where the data
is preprocessed by taking $64\times 64$ crops around each digit sequence. We
also use an additional more loosely $128\times 128$ cropped dataset as in
\cite{Ba14}. We train a baseline character sequence CNN model with 11 hidden
layers leading to five independent softmax classifiers, each one predicting the
digit at a particular position in the sequence. This is the character sequence
model used in \cite{Jaderberg14c}, where each classifier includes a null-character output to model variable length sequences. This model matches the
results obtained in \cite{Goodfellow13}.

We extend this baseline CNN to include a spatial transformer immediately
following the input (ST-CNN Single), where the localisation network is a
four-layer CNN. We also define another extension where before each of the first
four convolutional layers of the baseline CNN, we insert a spatial transformer
(ST-CNN Multi), where the localisation networks are all two layer fully
connected networks with 32 units per layer. In the ST-CNN Multi model, the
spatial transformer before the first convolutional layer acts on the input image
as with the previous experiments, however the subsequent spatial transformers
deeper in the network act on the convolutional feature maps, predicting a
transformation from them and transforming these feature maps (this is visualised
in \tblref{table:svhn} (right) (a)). This allows deeper spatial transformers to predict 
a transformation based on richer features rather than the raw image. All networks are trained from scratch with
SGD and dropout~\cite{Hinton12}, with randomly initialised weights, except for
the regression layers of spatial transformers which are initialised to predict
the identity transform. Affine transformations and bilinear sampling kernels are
used for all spatial transformer networks in these experiments.

The results of this experiment are shown in \tblref{table:svhn} (left) -- the
spatial transformer models obtain state-of-the-art results, reaching 3.6\% error
on $64 \times 64$ images compared to previous state-of-the-art of 3.9\% error.
Interestingly on $128\times 128$ images, while other methods degrade in
performance, an ST-CNN achieves 3.9\% error while the previous state of the art at
4.5\% error is with a recurrent attention model that uses an ensemble of models with
Monte Carlo averaging -- in contrast the ST-CNN models require only a single forward pass
of a single model. This accuracy is achieved due to the fact that the spatial
transformers crop and rescale the parts of the feature maps that correspond to
the digit, focussing resolution and network capacity only on these areas (see
\tblref{table:svhn} (right) (b) for some examples). In terms of computation speed, the ST-CNN Multi model is only 6\% slower (forward and backward pass) than the CNN.

\subsection{Fine-Grained Classification}\label{sec:fgc}

In this section, we use a spatial transformer network with multiple transformers in parallel to perform fine-grained bird classification. 
We evaluate our models on the CUB-200-2011 birds dataset~\cite{Wah11}, containing 6k training images and 5.8k test images, covering 200 species of birds. 
The birds appear at a range of scales and orientations, are not tightly cropped, and require detailed texture and shape analysis to distinguish.
In our experiments, we only use image class labels for training.

We consider a strong baseline CNN model -- an Inception architecture with batch normalisation~\cite{Ioffe15} pre-trained on ImageNet~\cite{Russakovsky14}
and fine-tuned on CUB -- which by itself achieves the state-of-the-art accuracy of 82.3\% (previous best result is 81.0\%~\cite{Simon15}). 
We then train a spatial transformer network, ST-CNN, which contains 2 or 4 parallel spatial transformers, parameterised for attention
and acting on the input image. Discriminative image parts, captured by the transformers, are passed to the part description sub-nets (each of which is
also initialised by Inception). The resulting part representations are concatenated and classified with a single softmax layer.
The whole architecture is trained on image class labels end-to-end with backpropagation (full details in Appendix~\ref{sec:morexp}).

The results are shown in \tblref{table:birds} (left).
The ST-CNN achieves an accuracy of 84.1\%, outperforming the baseline by 1.8\%. It should be noted that there is a small (22/5794) overlap between the ImageNet training set and CUB-200-2011 test set\footnote{Thanks to the eagle-eyed Hugo Larochelle and Yin Zheng  for spotting the birds nested in both the ImageNet training set and CUB test set.} -- removing these images from the test set results in 84.0\% accuracy with the same ST-CNN.
In the visualisations of the transforms predicted by {$2\times$ST-CNN} (\tblref{table:birds} (right)) one can see interesting behaviour has been learnt: 
one spatial transformer (red) has learnt to become a head detector, while the other (green) fixates on the central part of the body of a bird. 
The resulting output from the spatial transformers for the classification network is a somewhat pose-normalised representation of a bird. 
While previous work such as \cite{Branson14} explicitly define parts of the bird, training separate detectors for these parts with supplied 
keypoint training data, the ST-CNN is able to discover and learn part detectors in a data-driven manner without any additional supervision. In addition, the use of spatial transformers allows us to use 448px resolution input images without any impact in performance, as the output of the transformed 448px images are downsampled to 224px before being processed.

\begin{table}[t]
\begin{center}\small
\setlength{\tabcolsep}{3pt}
\begin{tabular}[t]{lr||c}
\multicolumn{2}{p{2.0cm}||}{\centering ~~~~Model} &
\multicolumn{1}{c}{\centering}\\
\hline
\multicolumn{2}{l||}{Cimpoi '15~\cite{Cimpoi15}} & 66.7\\
\multicolumn{2}{l||}{Zhang '14~\cite{Zhang14b}} & 74.9\\
\multicolumn{2}{l||}{Branson '14~\cite{Branson14}} & 75.7\\
\multicolumn{2}{l||}{Lin '15~\cite{Lin15}} & 80.9\\
\multicolumn{2}{l||}{Simon '15~\cite{Simon15}} & 81.0\\
CNN (ours) & 224px & 82.3\\
\hline
$2\times$ST-CNN & 224px & 83.1\\
$2\times$ST-CNN & 448px & 83.9\\
$4\times$ST-CNN & 448px & \bf{84.1}\\
\end{tabular}
\qquad\quad
\hspace{-2em}
\vtop{\vspace{-1em}\hbox{\includegraphics[width=0.7\textwidth]{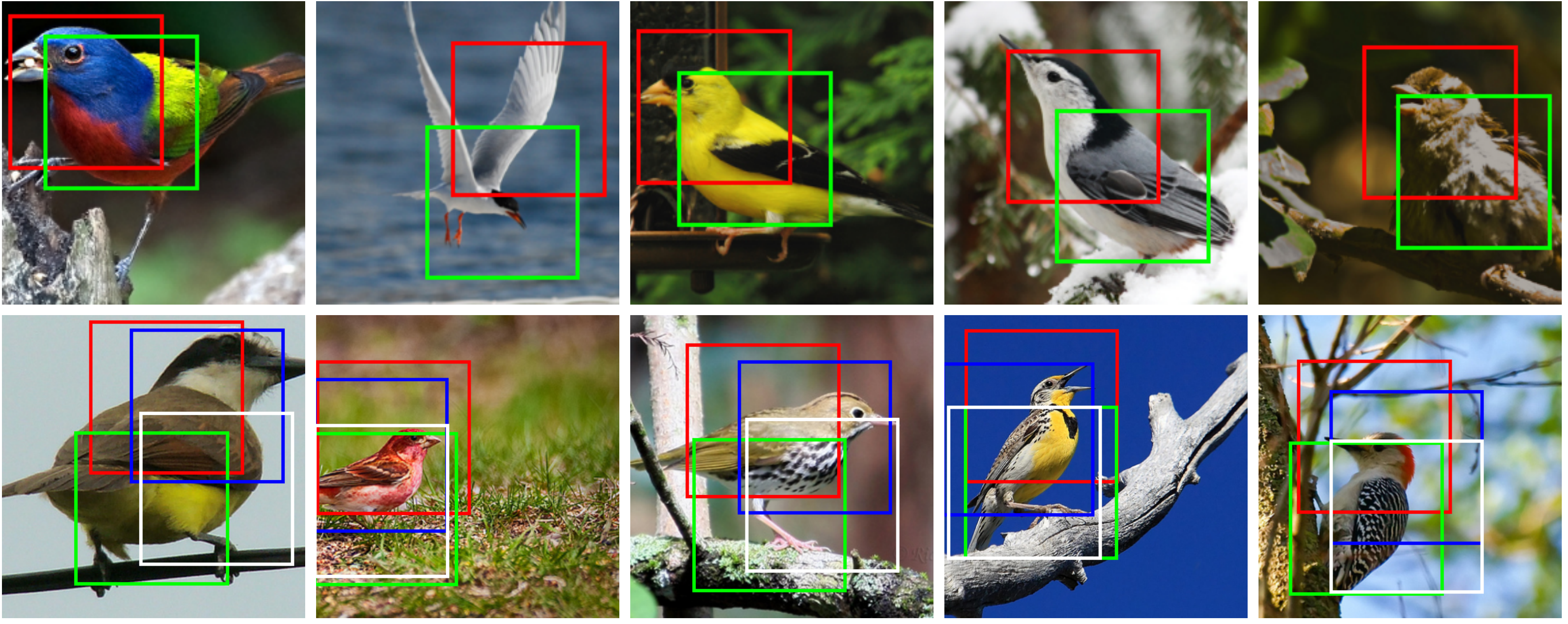}}}%
\end{center}
\vspace{-1em}
\caption{
\small \emph{Left:} The accuracy on CUB-200-2011 bird classification
dataset. Spatial transformer networks with two
spatial transformers ($2\times$ST-CNN) and four spatial transformers ($4\times$ST-CNN) in parallel achieve higher accuracy. 448px resolution images can be used with the ST-CNN without an increase in computational cost due to downsampling to 224px \emph{after} the transformers. \emph{Right:}
The transformation predicted by the spatial transformers of
$2\times$ST-CNN (top row)
 and $4\times$ST-CNN (bottom row)
on the input image. 
Notably for the $2\times$ST-CNN, one of the transformers (shown in red) learns to detect heads, 
while the other (shown in green) detects the body, and similarly for the $4\times$ST-CNN.
}
\label{table:birds}
\end{table}

\section{Conclusion}
In this paper we introduced a new self-contained module for neural
networks -- the spatial transformer. This module can be dropped into a
network and perform explicit spatial transformations of features,
opening up new ways for neural networks to model data, and is learnt
in an end-to-end fashion, without making any changes to the loss
function. While CNNs provide an incredibly strong baseline, we see
gains in accuracy using spatial transformers across multiple tasks,
resulting in state-of-the-art performance.  Furthermore, the regressed
transformation parameters from the spatial transformer are available
as an output and could be used for subsequent tasks.  
While we only
explore feed-forward networks in this work, early experiments show
spatial transformers to be powerful in recurrent models, and useful
for tasks requiring the disentangling of object reference frames, as well as easily extendable to 3D transformations (see Appendix~\ref{sec:3d}). 

\bibliographystyle{plainnat}
{
\small
\bibliography{shortstrings,vgg_local,vgg_other,current}
}

\begin{appendices}
\section{Appendix}\label{sec:morexp}
In this section we present the results of two further experiments -- that of MNIST addition showing spatial 
transformers acting on multiple objects in \sref{sec:mnistadd}, and co-localisation in \sref{sec:mnistcoloc} showing 
the application to semi-supervised scenarios. In addition, we give an example of the extension to 3D in~\sref{sec:3d}. We also expand upon the details of the experiments 
from \sref{sec:mnist} in \sref{sec:moremnist}, \sref{sec:svhn} in \sref{sec:moresvhn}, and \sref{sec:fgc} in \sref{sec:morefgc}.

\subsection{MNIST Addition}\label{sec:mnistadd}
\begin{table}[t]
\begin{center}\small
\setlength{\tabcolsep}{3pt}
\begin{tabular}[t]{ll||c}
\\
\\
\multicolumn{2}{p{2.0cm}||}{\centering ~~~~Model} &
\multicolumn{1}{c}{\centering RTS}\\
\hline
\multicolumn{2}{l||}{FCN} & 47.7\\
\multicolumn{2}{l||}{CNN} & 14.7\\
\hline
\multirow{3}{*}{ST-FCN}
    & Aff & 22.6\\
    & Proj & 18.5\\
    & TPS & 19.1\\
\hline
\multirow{3}{*}{$2\times$ST-FCN}
    & Aff & 9.0\\
    & Proj & 5.9\\
    & TPS & 5.8\\
\end{tabular}
\qquad\quad
\vtop{\vspace{-1em}\hbox{\includegraphics[width=0.65\textwidth]{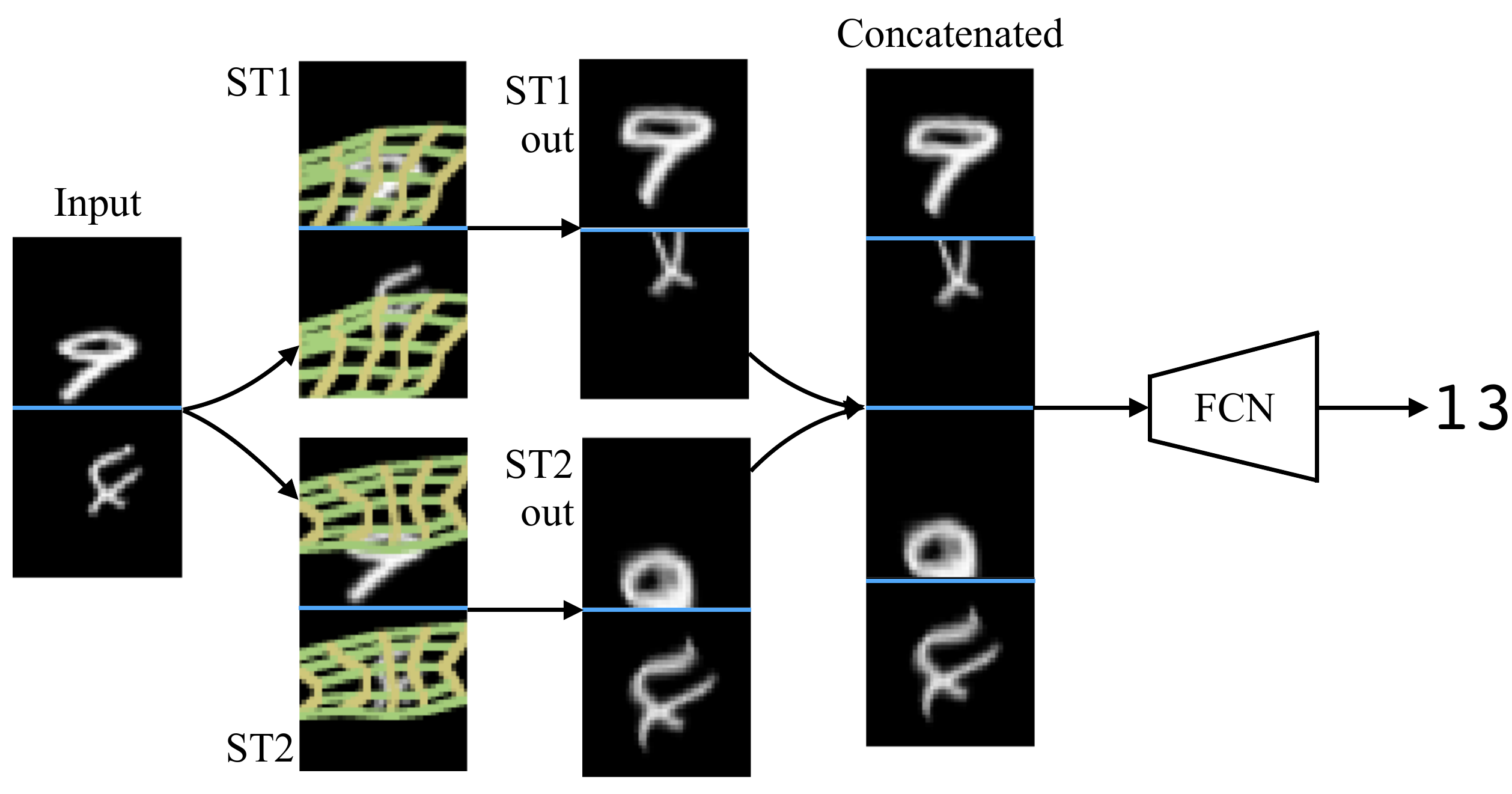}}}
\end{center}
\caption{\small \emph{Left:} The percentage error for the two digit MNIST addition task, where each
digit is transformed independently in separate channels, trained by supplying only the label of the sum of the two digits. The use of two spatial transformers in
parallel, $2\times$ST-FCN, allows the fully-connected neural network to become
invariant to the transformations of each digit, giving the lowest error. All the
models used for each column have approximately the same number of parameters. \emph{Right:} A test example showing the learnt behaviour of each spatial transformer (using a thin plate spline (TPS) transformation). The 2-channel input (the blue bar denotes separation between channels) is fed to two independent spatial transformers, ST1 and ST2, each of which operate on both channels. The outputs of ST1 and ST2 and concatenated and used as a 4-channel input to a fully connected network (FCN) which predicts the addition of the two original digits. During training, the two spatial transformers co-adapt to focus on a single channel each.}
\label{table:mnistadd}
\end{table}

In this section we demonstrate another use case for multiple spatial transformers in parallel: to model multiple objects. We define an MNIST addition task, where the network must output the sum of the two digits given in the input. Each digit is presented in a separate $42\times 42$ input channel (giving 2-channel inputs), but each digit is transformed independently, with random rotation, scale, and translation (RTS). 

We train fully connected (FCN), convolutional (CNN) and single spatial transformer fully connected (ST-FCN) networks, as well as spatial transformer fully connected networks with two parallel spatial transformers ($2\times$ST-FCN) acting on the input image, each one taking both channels as input and transforming both channels. The two 2-channel outputs of the two spatial transformers are concatenated into a 4-channel feature map for the subsequent FCN. As in \sref{sec:mnist}, all networks have the same number of parameters, and are all trained with SGD to minimise the multinomial cross entropy loss for 19 classes (the possible addition results 0-18).

The results are given in \tblref{table:mnistadd} (left). Due to the complexity of this task, the FCN reaches a minimum error of 47.7\%, however a CNN with max-pooling layers is far more accurate with 14.7\% error. Adding a single spatial transformer improves the capability of an FCN by focussing on a single region of the input containing both digits, reaching 18.5\% error. However, by using two spatial transformers, each transformer can learn to focus on transforming the digit in a single channel (though receiving both channels as input), visualised in \tblref{table:mnistadd} (right). The transformers co-adapt, producing stable representations of the two digits in two of the four output channels of the spatial transformers. This allows the $2\times$ST-FCN model to achieve 5.8\% error, far exceeding that of other models.

\subsection{Co-localisation}\label{sec:mnistcoloc}
\begin{table}[t]
\begin{center}\small
\setlength{\tabcolsep}{3pt}
\vspace{1em}
\begin{tabular}[t]{r||c|c}
\multicolumn{1}{c||}{~}&
\multicolumn{2}{c}{\centering MNIST Distortion}\\
Class & T & TC\\
\hline
0 & 100 & 81 \\
1 & 100 & 82 \\
2 & 100 & 88 \\
3 & 100 & 75 \\
4 & 100 & 94 \\
5 & 100 & 84 \\
6 & 100 & 93 \\
7 & 100 & 85 \\
8 & 100 & 89 \\
9 & 100 & 87 \\
\end{tabular}
\qquad\quad
\vtop{\vspace{-1em}\hbox{\includegraphics[width=0.6\textwidth]{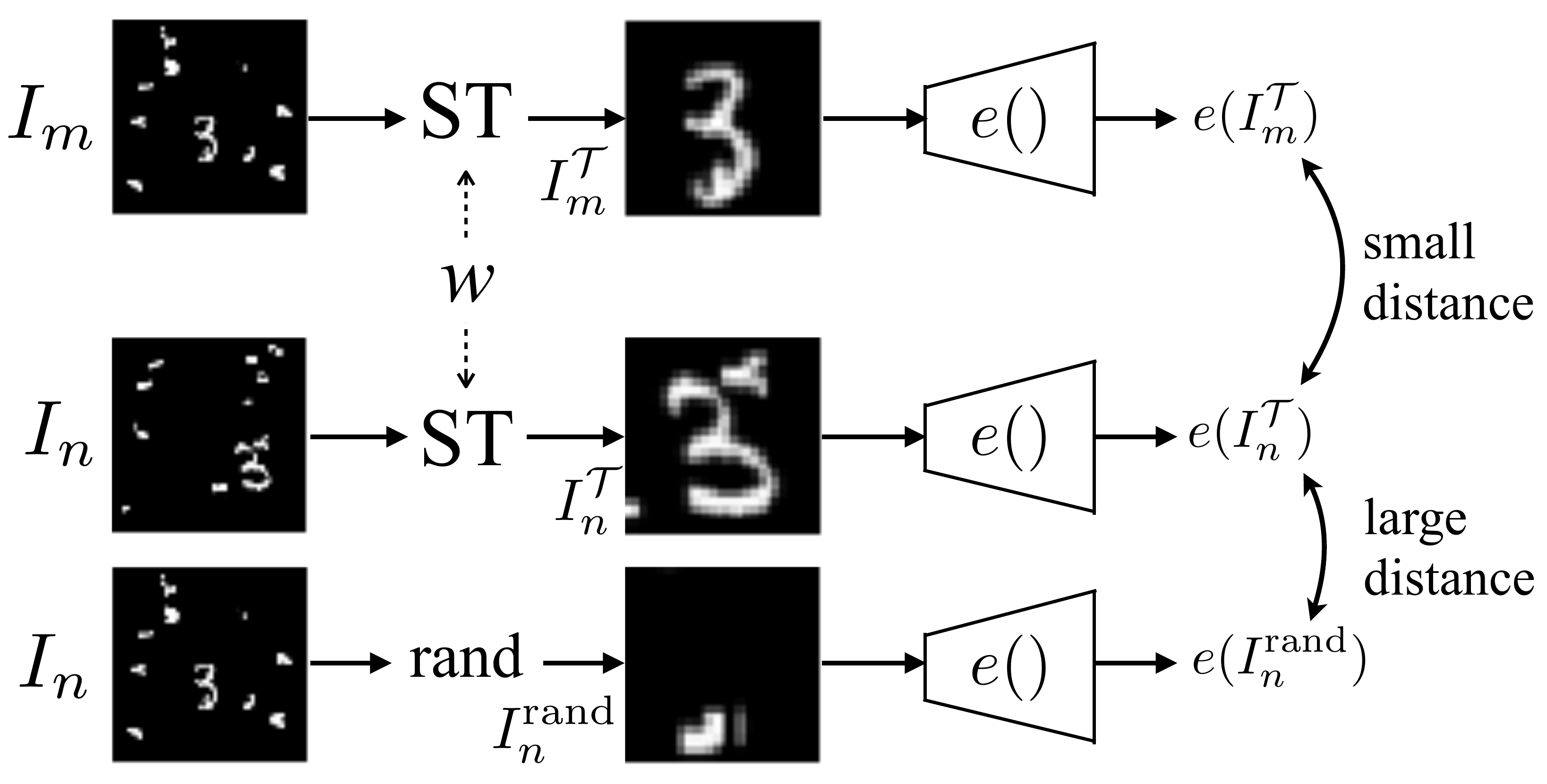}}}
\end{center}
\vspace{-1em}
\caption{\small \emph{Left:} The percent of correctly co-localised digits for different MNIST digit classes, for just translated digits (T), and for translated digits with clutter added (TC). \emph{Right:} The optimisation architecture. We use a hinge loss to enforce the distance between the two outputs of the spatial transformer (ST) to be less than the distance to a random crop, hoping to encourage the spatial transformer to localise the common objects.}
\label{table:coloc}
\end{table}

In this experiment, we explore the use of spatial transformers in a semi-supervised scenario -- co-localisation. The co-localisation task is as follows: given a set of images that are assumed to contain instances of a common but unknown object class, localise (with a bounding box) the common object. Neither the object class labels, nor the object location ground truth is used for optimisation, only the set of images.

To achieve this, we adopt the supervision that the distance between the image crop corresponding to two correctly localised objects is smaller than to a randomly sampled image crop, in some embedding space. For a dataset ${\mathcal I} = \{I_n\}$ of $N$ images, this translates to a triplet loss, where we minimise the hinge loss
\begin{equation}
\sum_{n}^{N} \sum_{m\ne n}^M \max(0, \|e(I_n^{\cal T}) - e(I_m^{\cal T})\|^2_2 - \|e(I_n^{\cal T}) - e(I_n^\text{rand})\|^2_2 + \alpha)
\label{eqn:hinge}
\end{equation}
where $I_n^{\cal T}$ is the image crop of $I_n$ corresponding to the localised object, $I_n^\text{rand}$ is a randomly sampled patch from $I_n$, $e()$ is an encoding function and $\alpha$ is a margin. We can use a spatial transformer to act as the localiser, such that $I_n^{\cal T} =  {\cal T}_\theta (I_n)$ where $\theta = f_\text{loc}(I_n)$, interpreting the parameters of the transformation $\theta$ as the bounding box of the object. We can minimise this with stochastic gradient descent, randomly sampling image pairs $(n,m)$.

We perform co-localisation on translated (T), and also translated and cluttered (TC) MNIST images. Each image, a $28\times 28$ pixel MNIST digit, is placed in a uniform random location in a $84\times 84$ black background image. For the cluttered dataset, we also then add 16 random $6\times 6$ crops sampled from the original MNIST training dataset, creating distractors. For a particular co-localisation optimisation, we pick a digit class and generate 100 distorted image samples as the dataset for the experiment. We use a margin $\alpha=1$, and for the encoding function $e()$ we use the CNN trained for digit classification from \sref{sec:mnist}, concatenating the three layers of activations (two hidden layers and the classification layer without softmax) to form a feature descriptor. We use a spatial transformer parameterised for attention (scale and translation) where the localisation network is a 100k parameter CNN consisting of a convolutional layer with eight $9\times 9$ filters and a 4 pixel stride, followed by $2\times 2$ max pooling with stride 2 and then two 8-unit fully-connected layers before the final 3-unit fully-connected layer.

\begin{figure}[t]
\centering
\begin{tabular}{cccc}
\includegraphics[width=0.8\textwidth]{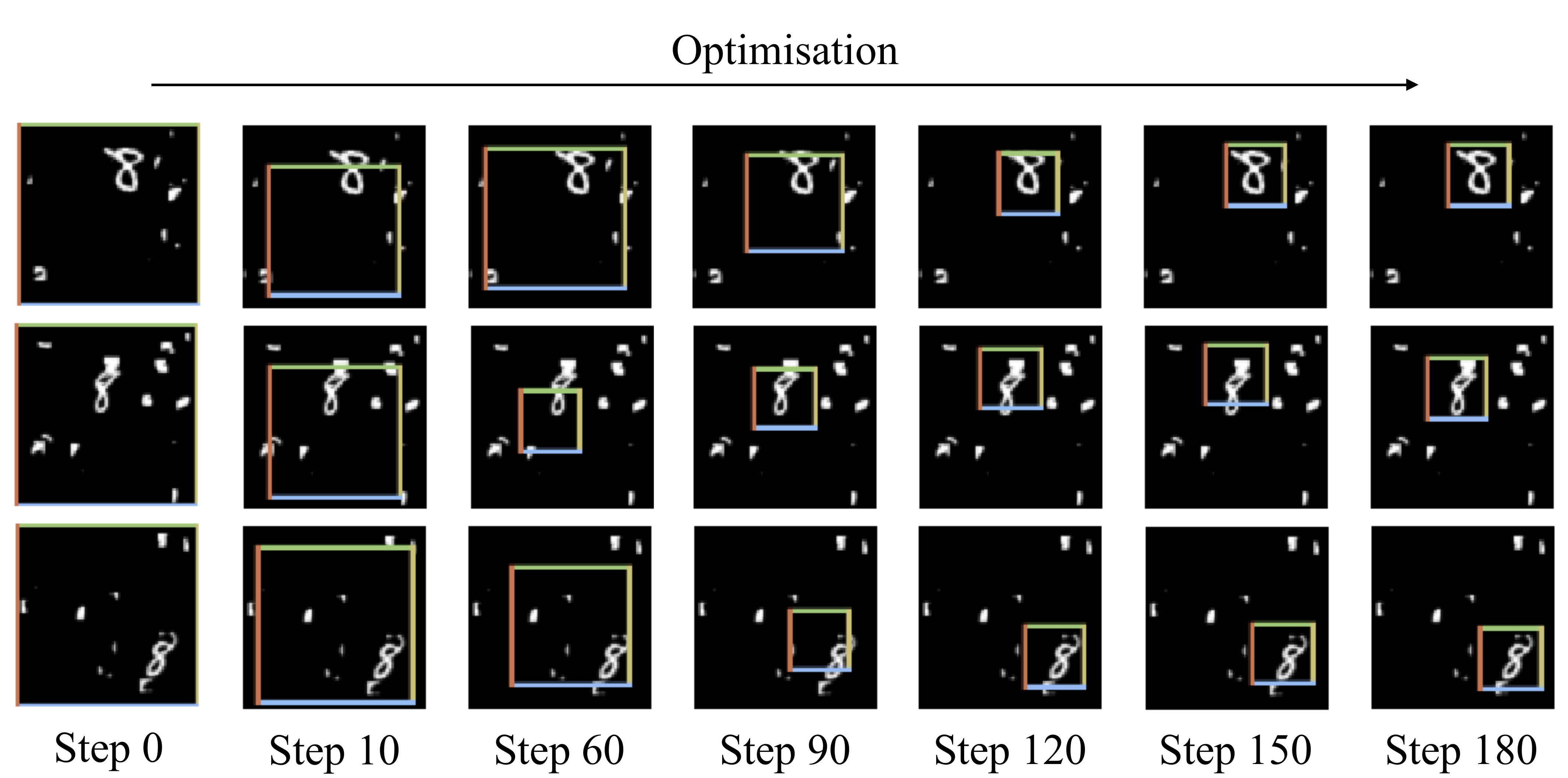}
\end{tabular}
\caption{\small A look at the optimisation dynamics for co-localisation. Here we show the localisation predicted by the spatial transformer for three of the 100 dataset images after the SGD step labelled below. By SGD step 180 the model has process has correctly localised the three digits. A full animation is shown in the video \url{https://goo.gl/qdEhUu}}
\label{fig:coloc}
\end{figure}

The results are shown in \tblref{table:coloc}. We measure a digit to be correctly localised if the overlap (area of intersection divided by area of union) between the predicted bounding box and groundtruth bounding box is greater than 0.5. Our co-localisation framework is able to perfectly localise MNIST digits without any clutter with 100\% accuracy, and correctly localises between 75-93\% of digits when there is clutter in the images. An example of the optimisation process on a subset of the dataset for ``8'' is shown in \figref{fig:coloc}. This is surprisingly good performance for what is a simple loss function derived from simple intuition, and hints at potential further applications in tracking problems.

\subsection{Higher Dimensional Transformers}
\label{sec:3d}
The framework described in this paper is not limited to 2D transformations and can be easily extended to higher dimensions. To demonstrate this, we give the example of a spatial transformer capable of performing 3D affine transformations. 

We extended the differentiable image sampling of \sref{sec:gridsampler} to perform 3D bilinear sampling. The 3D equivalent of \eqref{eqn:bilinear} becomes
\begin{equation}
V_i^c = \sum_{n}^{H} \sum_{m}^W \sum_{l}^D U^c_{nml}\max(0, 1-|x_i^s-m|)\max(0,1-|y_i^s-n|)\max(0,1-|z_i^s-l|)
\label{eqn:3Dbilinear}
\end{equation}
for the 3D input $U \in \mathbb{R}^{H \times W \times D \times C}$ and output $V \in \mathbb{R}^{H' \times W' \times D' \times C}$, where
$H'$, $W'$, and $D'$ are the height, width and depth of the grid, and $C$ is the number of channels. Similarly to the 2D sampling grid in \sref{sec:affinegrid}, the source coordinates that define the sampling points, $(x^s_i,y^s_i,z^s_i)$ can be generated by the transformation of a regular 3D grid $G=\{G_i\}$ of voxels $G_i = (x_i^t,y_i^t,z_i^t)$. For a 3D affine transformation this is
\begin{equation}
\left( {
\renewcommand{\arraystretch}{1.2}
\begin{array}{c}
x_i^s\\
y_i^s\\
z_i^s
\end{array}
}\right)
=
\left[ {
\renewcommand{\arraystretch}{1.2}
\begin{array}{cccc}
\theta_{11} & \theta_{12} & \theta_{13} & \theta_{14}\\
\theta_{21} & \theta_{22} & \theta_{23} & \theta_{24}\\
\theta_{31} & \theta_{32} & \theta_{33} & \theta_{34}\\
\end{array}
}\right]
\left( {
\renewcommand{\arraystretch}{1.2}
\begin{array}{c}
x^t_i\\
y^t_i\\
z^t_i\\
1
\end{array}
}\right).
\label{eqn:affine}
\end{equation}

The 3D spatial transformer can be used just like its 2D counterpart, being dropped into neural networks to provide a way to warp data in 3D space, where the third dimension could be space or time. 

\begin{figure}[t]
\centering
\begin{tabular}{cccc}
\includegraphics[width=0.8\textwidth]{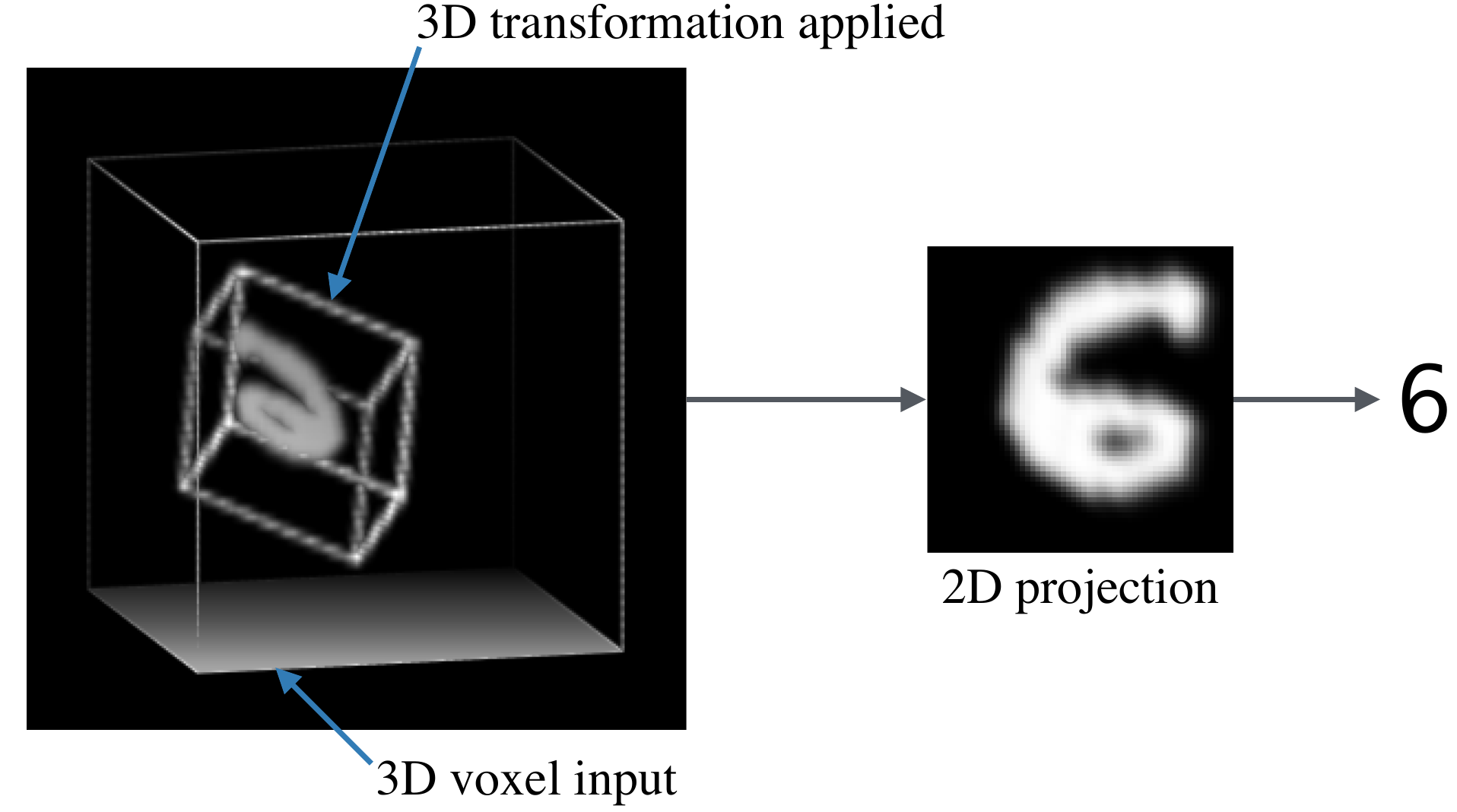}
\end{tabular}
\caption{\small The behaviour of a trained 3D MNIST classifier on a test example. The 3D voxel input contains a random MNIST digit which has been extruded and randomly placed inside a $60\times 60 \times 60$ volume. A 3D spatial transformer performs a transformation of the input, producing an output volume whose depth is then flattened. This creates a 2D projection of the 3D space, which the subsequent layers of the network are able to classify. The whole network is trained end-to-end with just classification labels.}
\label{fig:3dmnist}
\end{figure}

Another interesting way to use the 3D transformer is to flatten the 3D output across one dimension, creating a 2D projection of the 3D space, e.g. $W^c_{nm} = \sum_{l} V_{nml}^c$ such that $W \in \mathbb{R}^{H' \times W' \times C}$. This allows the original 3D data to be intelligently projected to 2D, greatly reducing the dimensionality and complexity of the subsequent processing. We demonstrated this on the task of 3D object classification on a dataset of 3D, extruded MNIST digits. The task is to take a 3D voxel input of a digit which has been randomly translated and rotated in 3D space, and output the class of the digit. The resulting 3D spatial transformer network learns to create a 2D projection of the 3D space where the digit is centered in the resulting 2D image, making it easy for the remaining layers to classify. An example is shown in~\figref{fig:3dmnist}.

\subsection{Distorted MNIST Details}\label{sec:moremnist}
In this section we expand upon the details of the distorted MNIST experiments in \sref{sec:mnist}. 

\paragraph{Data.} The rotated dataset (R) was generated from rotating MNIST training digits with a random rotation sampled uniformly between $-90^\circ$ and $+90^\circ$. The rotated, translated, and scaled dataset (RTS) was generated by randomly rotating an MNIST digit by $+45^\circ$ and $-45^\circ$, randomly scaling the digit by a factor of between 0.7 and 1.2, and placing the digit in a random location in a $42 \times 42$ image, all with uniform distributions. The projected dataset (P) was generated by scaling a digit randomly between 0.75 and 1.0, and stretching each corner of an MNIST digit by an amount sampled from a normal distribution with zero mean and 5 pixel standard deviation. The elasticly distorted dataset (E) was generated by scaling a digit randomly between 0.75 and 1.0, and then randomly peturbing 16 control points of a thin plate spline arranged in a regular grid on the image by an amount sampled from a normal distribution with zero mean and 1.5 pixel standard deviation. The translated and cluttered dataset (TC) is generated by placing an MNIST digit in a random location in a $60\times 60$ black canvas, and then inserting six randomly sampled $6\times 6$ patches of other digit images into random locations in the image.

\paragraph{Networks.} All networks use rectified linear non-linearities and softmax classifiers. All FCN networks have two hidden fully connected layers followed by a classification layer. All CNN networks have a $9\times 9$ convolutional layer (stride 1, no padding), a $2\times 2$ max-pooling layer with stride 2, a subsequent $7 \times 7$ convolutional layer (stride 1, no padding), and another $2\times 2$ max-pooling layer with stride 2 before the final classfication layer. All spatial transformer (ST) enabled networks place the ST modules at the beginning of the network, and have three hidden layers in their localisation networks with 32 unit fully connected layers for ST-FCN networks and two 20-filter $5\times 5$ convolutional layers (stride 1, no padding) acting on a $2\times$ downsampled input, with $2 \times 2$ max-pooling between convolutional layers, and a 20 unit fully connected layer following the convolutional layers. Spatial transformer networks for TC and RTS datasets have average pooling after the spatial transformer to downsample the output of the transformer by a factor of 2 for the classification network. The exact number of units in FCN and CNN based classification models varies so as to always ensure that all networks for a particular experiment contain the same number of learnable parameters (around 400k). This means that spatial transformer networks generally have less parameters in the classification networks due to the need for parameters in the localisation networks. The FCNs have between 128 and 256 units per layer, and the CNNs have between 32 and 64 filters per layer. 

\paragraph{Training.} All networks were trained with SGD for 150k iterations, the same hyperparameters (256 batch size, 0.01 base learning rate, no weight decay, no dropout), and same learning rate schedule (learning rate reduced by a factor of ten every 50k iterations). We initialise the network weights randomly, except for the final regression layer of localisation networks which are initialised to regress the identity transform (zero weights, identity transform bias). We perform three complete training runs for all models with different random seeds and report average accuracy.

\subsection{Street View House Numbers Details}\label{sec:moresvhn}
For the SVHN experiments in \sref{sec:svhn}, we follow \cite{Goodfellow13,Ba14} and select hyperparameters from a validation set of 5k images from the training set. All networks are trained for 400k iterations with SGD (128 batch size), using a base learning rate of 0.01 decreased by a factor of ten every 80k iterations, weight decay set to 0.0005, and dropout at 0.5 for all layers except the first convolutional layer and localisation networks. The learning rate for localisation networks of spatial transformer networks was set to a tenth of the base learning rate.

We adopt the notation that conv[$N$,$w$,$s$,$p$] denotes a convolutional layer with $N$ filters of size $w\times w$, with stride $s$ and $p$ pixel padding, fc[$N$] is a fully connected layer with $N$ units, and max[$s$] is a $s\times s$ max-pooling layer with stride $s$. The CNN model is: conv[48,5,1,2]-max[2]-conv[64,5,1,2]-conv[128,5,1,2]-max[2]-conv[160,5,1,2]-conv[192,5,1,2]-max[2]-conv[192,5,1,2]-conv[192,5,1,2]-max[2]-conv[192,5,1,2]-fc[3072]-fc[3072]-fc[3072], with rectified linear units following each weight layer, followed by five parallel fc[11] and softmax layers for classification (similar to that in \cite{Jaderberg14c}). The ST-CNN Single has a single spatial transformer (ST) before the first convolutional layer of the CNN model -- the ST's localisation network architecture is as follows: conv[32,5,1,2]-max[2]-conv[32,5,1,2]-fc[32]-fc[32]. The ST-CNN Multi has four spatial transformers, one before each of the first four convolutional layers of the CNN model, and each with a simple fc[32]-fc[32] localisation network.

We initialise the network weights randomly, except for the final regression layer of localisation networks which are initialised to regress the identity transform (zero weights, identity transform bias). We performed two full training runs with different random seeds and report the average accuracy obtained by a single model.

\subsection{Fine Grained Classification Details}\label{sec:morefgc}
\begin{figure}[t]
\centering
\begin{tabular}{cccc}
\includegraphics[width=0.8\textwidth]{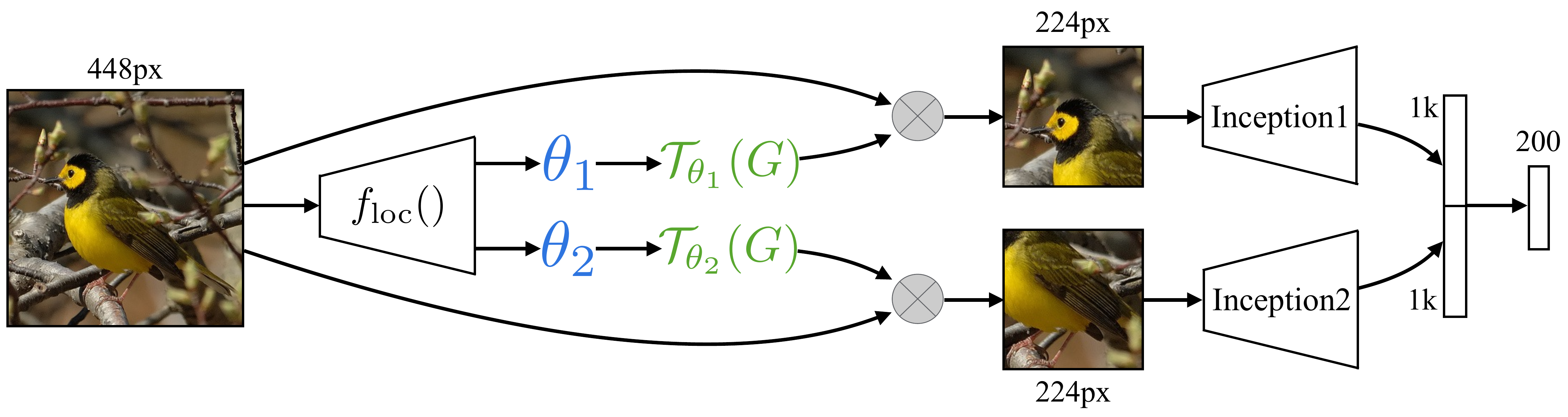}
\end{tabular}
\caption{\small The architecture of the $2\times$ST-CNN 448px used for bird classification. A single localisation network $f_\text{loc}$ predicts two transformation parameters $\theta_1$ and $\theta_2$, with the subsequent transforms ${\cal T}_{\theta_1}$ and ${\cal T}_{\theta_2}$ applied to the original input image.}
\label{fig:birdarch}
\end{figure}

In this section we describe our fine-grained image classification architecture in more detail.
For this task, we utilise the spatial transformers as a differentiable attention mechanism, where each transformer is expected to automatically learn to focus on discriminative object parts.
Namely, each transformer predicts the location (x,y) of the attention window, while the scale is fixed to $50\%$ of the image size.
The transformers sample $224 \times 224$ crops from the input image, each of which is then described each by its own CNN stream, thus forming a multi-stream architecture (shown in~\figref{fig:birdarch}).
The outputs of the streams are $1024$-D crop descriptors, which are concatenated and classified with a $200$-way softmax classifier.

As the main building block of our network, we utilise the state-of-the-art Inception architecture with batch normalisation~\cite{Ioffe15}, pre-trained on the ImageNet Challenge (ILSVRC) dataset.
Our model achieves $27.1\%$ top-1 error on the ILSVRC validation set using a single image crop (we only trained on single-scale images, resized so that the smallest side is $256$).
The crop description networks employ the Inception architecture with the last layer ($1000$-way ILSVRC classifier) removed, so that the output is a $1024$-D descriptor.

The localisation network is shared across \emph{all} the transformers, and was derived from Inception in the following way. Apart from the ILSVRC classification layer, we also removed the last pooling layer
to preserve the spatial information. The output of this truncated Inception net has $7 \times 7$ spatial resolution and $1024$ feature channels.
On top of it, we added three weight layers to predict the transformations: 
(i) $1\times1$ convolutional layer to reduce the number of feature channels from $1024$ to $128$; 
(ii) fully-connected layer with $128$-D output;
(iii) fully-connected layer with $2N$-D output, where $N$ is the number of transformers (we experimented with $N=2$ and $N=4$). 

We note that we did not strive to optimise the architecture in terms of the number of parameters and the computation time. 
Our aim was to investigate whether spatial transformer networks are able to automatically discover meaningful object parts when trained just on image labels, which
we confirmed both quantitatively and qualitatively (\sref{sec:fgc}).

The model was trained for 30k iterations with SGD (batch size 256) with an initial learning rate of 0.1, reduced by a factor of 10 after 10k, 20k, and 25k iterations. For stability, the localisation network's learning rate is the base learning rate multiplied by $10^{-4}$. Weight decay was set at $10^-5$ and dropout of 0.7 was used before the 200-way classification layer.

We evaluated two input images sizes for the spatial transformers: $224\times 224$ and $448\times 448$. In the latter case, we added a fixed $2\times$ downscaling layer before the localisation net,
so that its input is still $224\times 224$. The difference between the two settings lies in the size of the image from which sampling is performed ($224$ vs $448$), with $448$ better
suited for sampling small-scale crops. The output of the transformers are $224\times224$ crops in both cases (so that they are compatible with crop description Inception nets).
When training, we utilised conventional augmentation in the form of random sampling ($224\times 224$ from $256\times S$ and $448\times 448$ from $512\times S$ where $S$ is the largest image side) and horizontal flipping.
The localisation net was initialised to tile the image plane with the spatial transformer crops.

We also experimented with more complex transformations (location and scale, as well as affine), but observed similar results.
This can be attributed to the very small size of the training set (6k images, 200 classes), and we noticed severe over-fitting in all training scenarios.
The hyper-parameters were estimated by cross-validation on the training set.

\end{appendices}

\end{document}